\documentclass[10pt,twocolumn,letterpaper]{article}

\usepackage[pagenumbers]{iccv} 

\definecolor{iccvblue}{rgb}{0.21,0.49,0.74}
\usepackage[pagebackref,breaklinks,colorlinks,allcolors=iccvblue]{hyperref}

\usepackage{textcomp}
\usepackage{stfloats}
\usepackage{url}
\usepackage{verbatim}
\usepackage{graphicx}
\usepackage{multirow}
\usepackage{booktabs}
\usepackage{caption}
\usepackage{graphicx}
\usepackage{float} 
\usepackage{doi}
\usepackage{textcomp}
\usepackage{colortbl}  
\usepackage{xcolor}
\usepackage{array}   

\def\paperID{7094} 
\def\confName{ICCV}
\def\confYear{2025}

\title{From Image- to Pixel-level: Label-efficient Hyperspectral Image Reconstruction}

\author{Yihong~Leng\\
{\tt\small yhleng@stu.xidian.edu.cn}
\and
Jiaojiao~Li\\
{\tt\small jjli@xidian.edu.cn}
\and
Haitao~XU\\
{\tt\small xuhaitao@nssc.ac.cn}
\and
Rui~Song\\
{\tt\small rsong@xidian.edu.cn}
}

\begin{document}
\maketitle

\begin{abstract}
Current hyperspectral image (HSI) reconstruction methods primarily rely on image-level approaches, which are time-consuming to form abundant high-quality HSIs through imagers. 
In contrast, spectrometers offer a more efficient alternative by capturing high-fidelity point spectra, enabling pixel-level HSI reconstruction that balances accuracy and label efficiency.
To this end, we introduce a pixel-level spectral super-resolution (Pixel-SSR) paradigm that reconstructs HSI from RGB and point spectra.
Despite its advantages, Pixel-SSR presents two key challenges:
1) generalizability to novel scenes lacking point spectra, and 2) effective information extraction to promote reconstruction accuracy.
To address the first challenge, a Gamma-modeled strategy is investigated to synthesize point spectra based on their intrinsic properties, including nonnegativity, a skewed distribution, and a positive correlation. 
Furthermore, complementary three-branch prompts from RGB and point spectra are extracted with a Dynamic Prompt Mamba (DyPro-Mamba), which progressively directs the reconstruction with global spatial distributions, edge details, and spectral dependency.
Comprehensive evaluations, including horizontal comparisons with leading methods and vertical assessments across unsupervised and image-level supervised paradigms, demonstrate that ours achieves competitive reconstruction accuracy with efficient label consumption.

\end{abstract}    
\section{Introduction}
\label{sec:intro}
Hyperspectral image (HSI) offers significant advantages in identifying material compositions across a wide range of contiguous and narrow spectral bands, finding applications in object detection\cite{dong2024hypertdTCYB}, classification \cite{Zhao_2023_ICCV, li2024swformerTIP}, remote sensing \cite{Zeng_2024_CVPR, li2024testTPAMI}, and medical image analysis \cite{ng2024hyperNips} etc.
Nonetheless, acquiring high-quality HSIs entails heavy photon energy consumption and prolonged exposure times, posing stringent requirements on imaging hardware, environmental conditions, and temporal constraints in real-world scenes.

To address these limitations, researchers have explored various computational strategies for HSI reconstruction from degraded HSIs, low-dimensional RGBs or multi-spectral images(MSIs), compressed 2D measurements via coded aperture snapshot spectral imaging (CASSI), or low-resolution HSIs with RGB/MSI, etc.
Traditional methods typically impose handcrafted priors to model spatial and spectral degradation, such as sparse representation \cite{akhtar2015bayesianCVPR, arad2016sparseECCV} and matrix decomposition \cite{zhang2013MatrixTGRS, dian2017hyperspectralCVPR}.
However, these priors, predefined under specific assumptions, lack robustness and adaptability in complex imaging environments.

\begin{figure}[!t]
	\centering
	\includegraphics[width=3.3in]{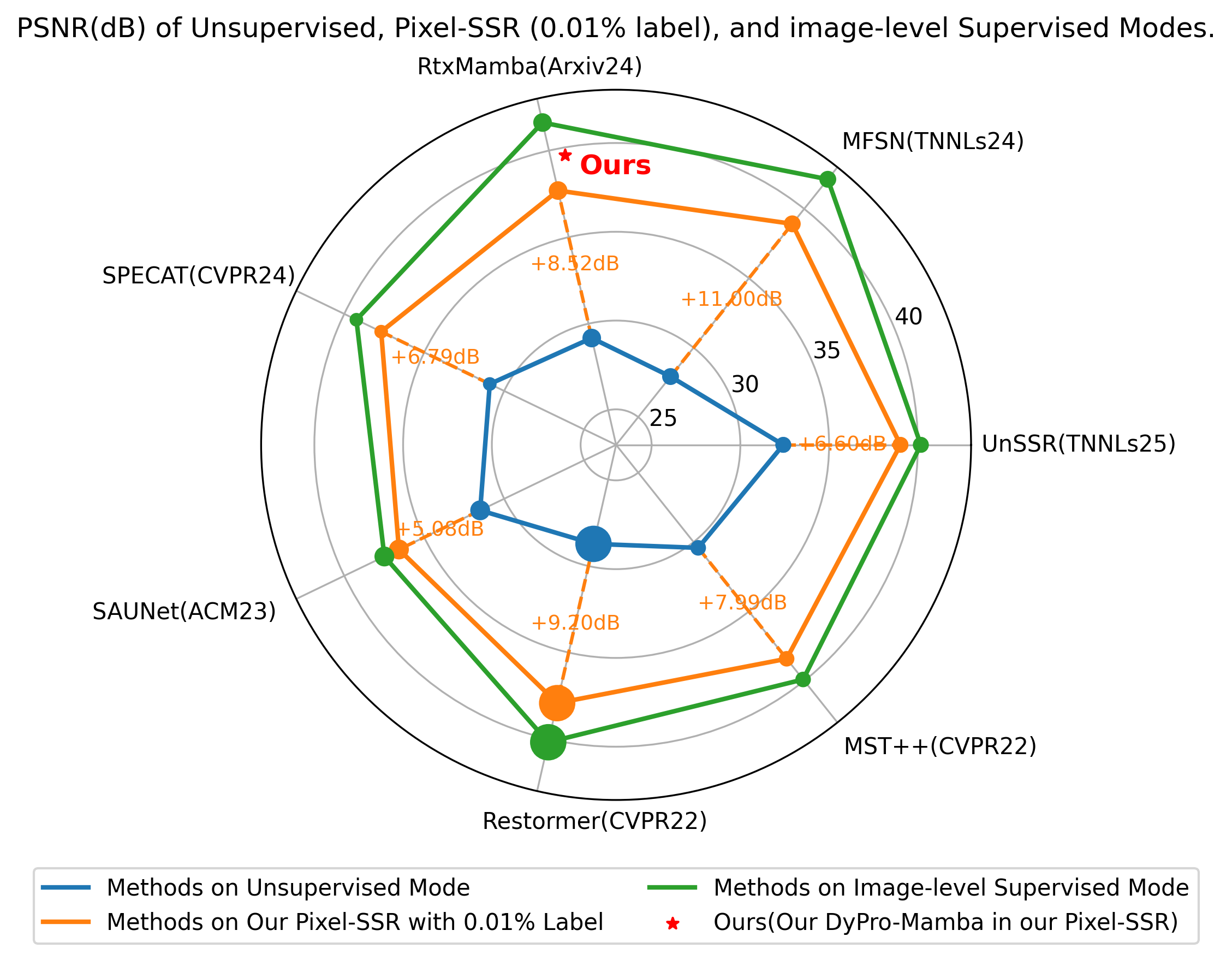}%
    \caption{\textbf{Reconstruction Performance Comparisons.} Blue, Orange, and Green denote SOTA methods in unsupervised, our proposed Pixel-SSR, and image-level supervised modes. Advances from Blue to Orange demonstrate that our universal Pixel-SSR significantly enhances the precision with efficient 0.01\% labels. Ours, marked as a red star, achieves the best results among all methods in Pixel-SSR mode (Orange) and comparable results to those in image-level supervised mode (Green).  Marker sizes reflect the methods' parameter counts.}
    \label{fig1}
\end{figure}

Recently, deep learning-based methods \cite{shi2018hscnnCVPR, guo2022deeppami, guo2022cmtCVPR, huo2024SSRTIP} have demonstrated superior capability in feature representation and have ushered in new techniques for HSI reconstruction. 
In particular, non-local spatial \cite{AWAN, restormerCVPR2022} and global spectral attention mechanisms \cite{cai2022mst++, yang2024hyperspectralAAAI} have proven pivotal in capturing rich spatial patterns and intrinsic spectral dependencies.
Despite these benefits, scalability remains challenging due to quadratic complexity when processing images with wide widths and high channels.
More recently, Mamba \cite{gu2023mamba} has emerged as a scalable alternative, modeling long-range dependencies with linear complexity through a state space model (SSM).  It has been successfully applied in vision tasks \cite{zhu2024visionmamba, zhen2024freqmamba} and HSI classification domain \cite{li2024mambahsiTGRS, yao2024spectralmamba}.

Nevertheless, supervised HSI reconstruction methods remain predominantly image-level, which heavily rely on large and data-driven networks, often resulting in reduced model transferability and robustness.
In contrast, unsupervised approaches, despite eliminating reliance on labeled data, often suffer from accuracy and spectral fidelity limitations. 
Rather than collecting the entire images, spectrometers provide a practical alternative by capturing high-resolution point spectra with minimal cost and rapid acquisition speed. 
This motivates us to explore a Pixel-Level Spectral Super-Resolution (Pixel-SSR) paradigm, which reconstructs HSIs from RGB images and sparse point spectra, offering a label-efficient alternative that balances accuracy with practicality.
Despite its advantages, Pixel-SSR introduces two fundamental challenges:
\textbf{1) Generalizability to New Scenes:} How can the model retain accuracy when applied to novel scenes where point spectra are unavailable for fine-tuning or adaptation?
\textbf{2) Enhancing Information Extraction:} From image- to pixel-level constraints, how can we effectively mine and then leverage latent priors to guide a precise and robust reconstruction? 

To address the first challenge, substituting point spectra with a specific distribution serves as a spontaneous solution in scenes without point spectra.
We analyze the properties of point spectra that exhibit nonnegativity, a skewed distribution, and a positive correlation.
Thus, the Gamma distribution is investigated to model point spectra during both training and inference, providing a fresh perspective to enhance the model's portability and robustness.
For the second challenge, it is evident that RGB boasts abundant spatial information yet lacks spectral details, whereas few point spectra offer precise spectral attributes but deficient spatial distribution.
Inspired by their complementary nature, a Dynamic Receptive Prompt Neck (DRPN) with a three-branch structure, is proposed to extract Spa-FRFT Prompt $\mathcal{P}_{spa}$, Spa-HF Prompt $\mathcal{P}_{hf}$, and Spectral Prompt $\mathcal{P}_{spe}$.
\begin{itemize}
	\item $\mathcal{P}_{spa}$: Captures non-stationary content through Fractional Fourier Transform (FRFT), depicting global spatial distributions in a hybrid-sweep way.
    \item $\mathcal{P}_{hf}$: Enhances high-frequency details to excavate spatial texture and edge clarity.
    \item $\mathcal{P}_{spe}$: Integrates outputs from $\mathcal{P}_{spa}$ and $\mathcal{P}_{hf}$, injects spatial detail into the Gamma-modeled point spectra, and delineates spectral correlations through self-attention.
\end{itemize}
Next, PromptSSM is presented to fuse the previously extracted prompts, capturing long-range spatio-spectral dependencies via leveraging a spectral-prompt mechanism to guide the spatial SSM.
Finally, DRPN with PromptSSM forms our robust and versatile Dynamic Prompt Mamba (DyPro-Mamba), which achieves competitive results compared to leading methods, as illustrated in Fig.\ref{fig1}.

Our main contributions are summarized as follows:
\begin{enumerate}
    \item{We take the lead in the Pixel-SSR task, providing a label-efficient reconstruction paradigm to address practical label constraints while maintaining accuracy. Exhaustive experiments on three benchmarks demonstrate the portability, superiority, and robustness of our method.}
    \item{ 
    Gamma-modeled point spectra provide a feasible and robust solution to employ Pixel-SSR in new scenes.
    }
    \item{DyPro-Mamba is presented to dynamically excavate latent prompts $\mathcal{P}_{spa}$, $\mathcal{P}_{hf}$, and $\mathcal{P}_{spe}$ with DRPN, 
    followed by PromptSSM integration to refine spatio-spectral characteristics of reconstructed HSI.
    }
\end{enumerate}


\section{Related Work}
\label{sec: related work}

\subsection{Hyperspectral Image Reconstruction}
Deep learning has revolutionized HSI reconstruction, offering computational strategies capable of adapting to diverse imagers. Below, we categorize existing approaches based on the source data used for HSI reconstruction.

Reconstruction from degraded HSIs affected by spatial blur, noise, or atmospheric conditions.
Early work by \cite{he2020HSIResTPAMI} introduced a unified restoration network that achieved superior performance through a non-local low-rank denoising strategy.
Only with degraded HSI, \cite{miao2023dds2mICCV} presented a self-supervised diffusion model that infers restoration parameters during the reverse diffusion process.
Building upon a pre-trained diffusion model, \cite{pang2024hirCVPR} broke down the restoration process into two low-rank components to represent the reduced image and the coefficient matrix, improving the unsupervised restoration performance. 

Reconstruction from low-dimensional RGB or MSI: spectral reconstruction or spectral super-resolution(SSR).
NTIRE Spectral Reconstruction Challenge in 2018, 2020, and 2022 has driven innovations in recovery accuracy, focusing on convolutional architectures \cite{zhao2020hrnet}, imaging parameters integration \cite{AWAN}, and attention mechanisms \cite{cai2022mst++}.
To enhance accuracy in specific spectral bands, \cite{yang2024hyperspectralAAAI} proposed a separate reconstruction approach for Blue, Red, and Green bands.
Considering the practical limitation, unsupervised paradigms have been progressively proposed to customize powerful constraints from imaging process \cite{li2023mformer}, semantic consistency\cite{zhu2021semantic}, and uncertainty estimation\cite{leng2025uncertaintyTnnls}.

Reconstruction from a compressed 2D image by coded aperture snapshot spectral imaging (CASSI). 
Several researchers formulated this problem as an end-to-end regression process with customized self-attention\cite{mst}, frequency-level refinement\cite{hdnet}, and efficient network design\cite{yao2024specatCVPR}.
To address the lack of interpretability, deep unfolding algorithms have been employed to embed the priors of the CASSI signal encoding process \cite{wang2023saunetACM, li2023pixelICCV} into networks, promoting the performance and interpretability. 

Reconstruction from both low-dimensional RGB/MSI and low-resolution HSI: fusion-based methods.
\cite{xie2020mhfTPAMI, guo2022deepTPAMI, qu2024s2cyclediffAAAI} reconstructed the high-quality HSI via fitting the degradation process in spatial and spectral dimensions with customized optimization functions.
To better utilize available data, spectral unmixing has been integrated into fusion frameworks \cite{yu2024unmixingCVPR}, extracting spatial distributions and learnable abundance information to improve fusion performance.

Given the scarcity of publicly available HSI datasets compared to RGB images, SSR has become a crucial approach for HSI reconstruction. 
However, supervised SSR methods require abundant labeled HSIs, which contradicts our goal of minimizing supervision, while existing unsupervised SSR methods still suffer from accuracy and fidelity limitations. 
To bridge this gap, we explore the Pixel-SSR paradigm, leveraging RGB images and sparse point spectra to reconstruct high-quality HSIs.

\subsection{State Space Model}
State Space Model (SSM) has demonstrated fabulous capability in modeling long-range dependencies with linear scalability.
It leverages a hidden state $h(t) \in R^{L} $ to map a 1D input $x(t) \in R^{L}$ to an output signal $y(t) \in R^{L}$:
\begin{equation}
\left\{\begin{aligned}
h^{\prime}(t) & =\mathbf{A} h(t)+\mathbf{B} x(t) \\
y(t) & =\mathbf{C} h(t)+\mathbf{D} x(t)
\end{aligned}\right.
\end{equation}
where $\mathbf{A} \in R^{N \times N}$, $\mathbf{B} \in R^{N \times 1}$ and $\mathbf{C} \in R^{1 \times N}$ are learnable parameters, and $\mathbf{D} \in R^{1}$ represents a residual connection.
Then, S4 \cite{gu2021efficiently} introduced deep learning into SSM, thus improving its performance with stability and affordable training speeds.
Furthermore, Mamba \cite{gu2023mamba} emerged as a standout work with a selective mechanism and efficient hardware acceleration, demonstrating superior performance to CNN and Transformer. 
Several Mamba-based works have showcased fabulous performance in image restoration \cite{bai2024retinexmamba, zhen2024freqmamba}, segmentation \cite{ruan2024vm, xing2024segmamba}, and hyperspectral image classification \cite{li2024mambahsiTGRS, yao2024spectralmamba, sheng2024dualmamba}.

\section{Methods}
\label{sec:methods}
Towards Pixel-SSR, we introduce a Gamma-modeled strategy for robust adaption to new scenes and DyPro-Mamba for enhancing information extraction in turn, jointly for label-efficient high-quality HSI reconstruction. 

\begin{figure}[!t]
	\centering
    \includegraphics[width=3.0in]{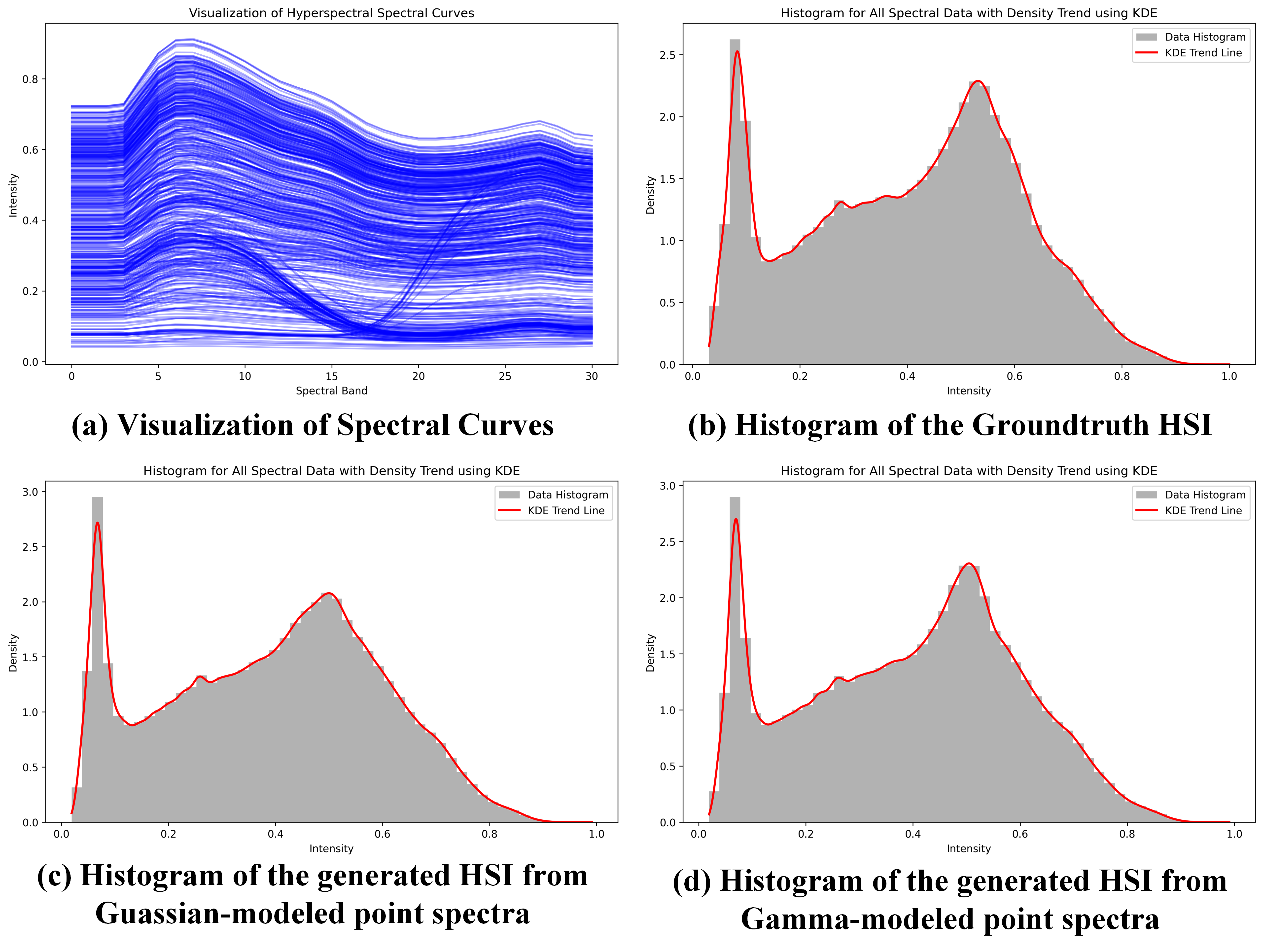}%
    \caption{ \textbf{Visualizations.}  The spectral curves  (a) and the histogram of the groundtruth HSI (b) both demonstrate skewness. The reconstructed HSI via Gamma-modeled (c) achieves a more precise distribution than via Gaussian-modeled (d).}
    \label{gamma}
\end{figure}


\subsection{Gamma-modeled Point Spectra Generation.} 
The primary challenge is to ensure the adaptability of Pixel-SSR in novel scenes where point spectra are unavailable. 
A straight solution would be to form point spectra with a specific distribution during inference. 
We visualize the spectra curves and data distribution in Fig. \ref{gamma}, revealing:
\begin{itemize} 
\item Nonnegativity: HSI values, representing radiation or reflectance, are inherently nonnegative.
\item Skewed distribution: Both the spectral curves and histograms exhibit skewness, which may stem from the presence of distinct absorption peaks in various materials.
\item Positive correlation: As HSI measurements originate from photon counts accumulated over finite time intervals, spectral values exhibit positive temporal correlation.
\end{itemize}
To address these features, especially skewness, we model point spectra with Gamma distribution.

The detailed processes can be described as: 
First, an initial Gamma-based distribution is formulated as $\mathbf{Y}_{g} \in R^{B \times H \times W} \sim \Gamma(\alpha, \beta)$, where $\alpha$ and $\beta$ individually denote the shape and scale parameters.
Then, a dynamic mask $\mathcal{M} \in R^{B \times H \times W} = True$ is designed to adjust the ratio of point spectra: 
\begin{equation}
     \mathcal{M}{\{B,i,j\}} = False, i \times j = \omega \times H \times W
\end{equation}
in which the spatial index (i,j) is randomly selected and $\omega$ stands for the ratio. 
Next, the final Gamma-modeled point spectra $\mathbf{Y}_{g}^{p}$ can be generated via $\mathbf{Y}_{g}^{p} = \mathbf{Y}_{g} \otimes \mathcal{M}$.

At last, $\mathbf{Y}_{g}^{p}$ consistently throughout the model process, which can be formulated as:
\begin{equation}
\begin{aligned}
    & \text{Training:}  \{\mathbf{Y}_{g}^{p}, \mathbf{X} \} \rightarrow \mathbf{Y}^{r}, \mathcal{L}(\mathbf{Y}^p, \mathbf{Y}^r)  \\
    & \text{Testing:} \{\mathbf{Y}_{g}^{p}, \mathbf{X} \} \rightarrow \mathbf{Y}^{r}
\end{aligned}
\end{equation}
where $\mathbf{X}$, $\mathbf{Y}^{r}$, $\mathbf{Y}^{p}$ denote RGB, reconstructed HSI and the groundtruth point spectra.
In this way, our trained Pixel-SSR reconstructs high-quality HSIs with Gamma-modeled point spectra and achieves an implementation in scenes without point spectra. Besides, the constraints $\mathcal{L}(\mathbf{Y}^p, \mathbf{Y}^r)$ is formulated in Supplementary.

\begin{figure*}[!t]
	\centering
	\includegraphics[width=0.95\textwidth]{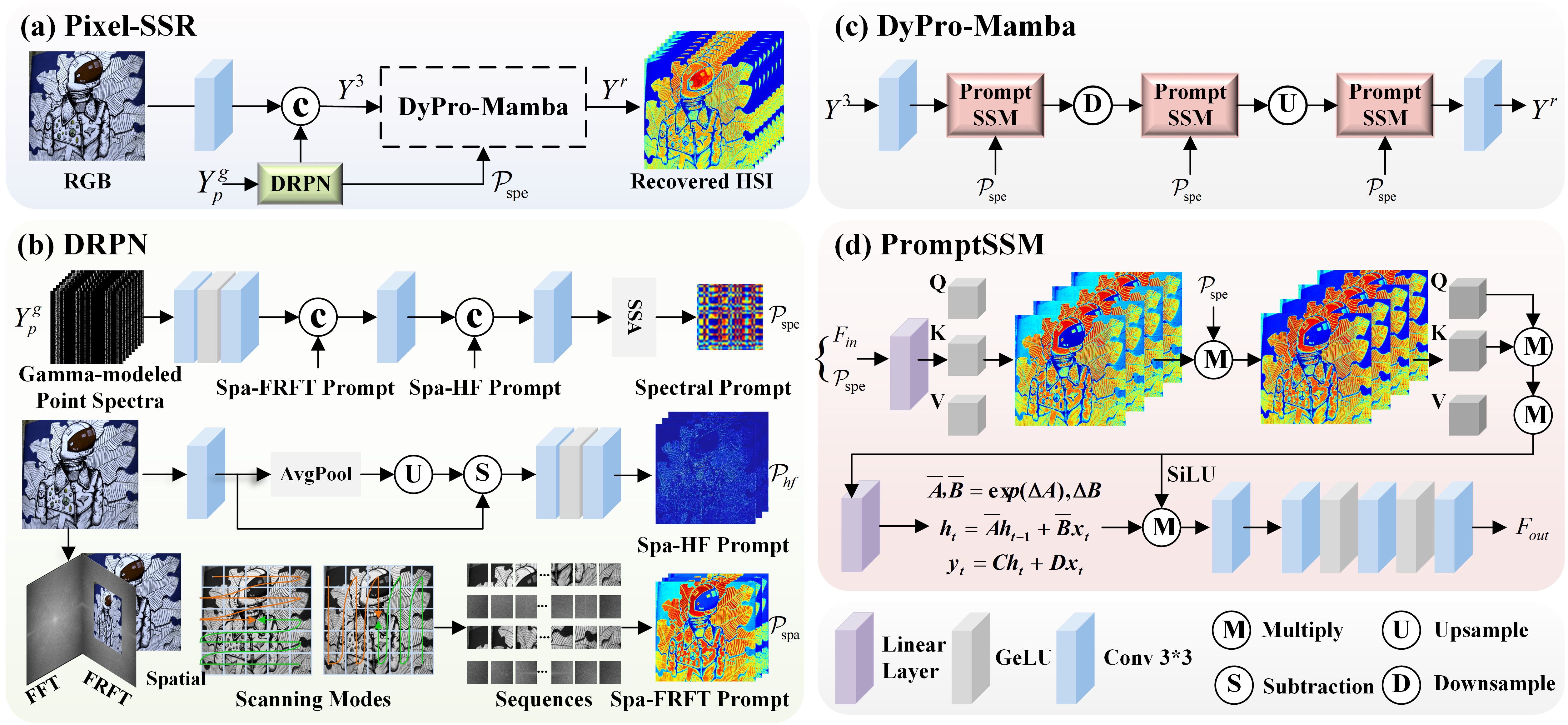}%
	\caption{ \textbf{Overview of our Pixel-SSR paradigm.} At its core, the \textbf{D}ynamic \textbf{R}eceptive \textbf{P}rompt \textbf{N}eck (DRPN) is designed to represent Spa-FRFT Prompt, Spa-HF Prompt, and Spectral Prompt, respectively denoted as spatial-wise sequences within the Fractional Fourier Transform (FRFT) domain, high-frequency representation, and spectral-wise dependency. These multi-type features undergo our \textbf{Dy}namic \textbf{Pro}mpt Mamba (DyPro-Mamba) with several PromptSSMs to delineate high-quality HSIs. When employed as a universal Pixel-SSR paradigm, the dotted box in (a) can be replaced with various leading methods, sharing the same DRPN and loss constraints with Ours.}
	\label{framework}
\end{figure*}

\subsection{Pixel-SSR.}
For the second challenge, a novel DyPro-Mamba is presented for Pixel-SSR, with a Dynamic Receptive Prompt Neck(DRPN) to extract inherent prompts and several PromptSSMs to reconstruct high-quality HSIs.  As illustrated in Fig.\ref{framework}, the overall pipeline can be implemented as a universal paradigm, replacing the dotted box with various leading methods when applied in practice. Detailed experimental verifications can be found in Section \ref{universal paradigm}.

\subsubsection{Dynamic Receptive Prompt Neck(DRPN).}
With the point spectra $\mathbf{Y}_{g}^{p}$ and the RGB $\mathbf{X} \in R^{b \times H \times W} $, DRPN is designed to excavate the inherent prompt representation from $\mathbf{X}$ and $\mathbf{Y}_{g}^{p}$.
In specific, $\mathbf{X}$ contains abundant spatial distributions while $\mathbf{Y}_{g}^{p}$ contributes informant spectral characteristics.
Thus, three independent branches are customized to model abundant spatial-wise details within the Fractional Fourier Transform (FRFT) domain (Spa-FRFT Prompt), high-frequency representation along the spatial dimension (Spa-HF Prompt), and spectral-wise dependency (Spectral Prompt).

\textbf{Spa-FRFT Prompt $\mathcal{P}_{spa}$}:
Natural images are non-stationary signals as their statistical properties vary with many factors, e.g., content, color, lighting parameters, and imaging conditions.
Besides, the non-stationary content in an image contains textures, edge detail, spectral absorption peaks, etc.
Thus, FRFT\cite{yu2024FRFTNIPs} is leveraged to process the non-stationary content for preserving more image details than the Fast Fourier Transform (FFT).
Furthermore, VisionMamba\cite{zhu2024visionmamba} demonstrates outperforming capability in modeling global spatial features with linear complexity.
However, the spatial relationship is more complex than a sequence in NLP, resulting in an incomplete location modeling.
Thus, the strategy to transform the 2D image to 1D sequences partly determines how much spatial relations can be preserved in 1D sequences.
Various methods have been researched to increase scanning routes \cite{zhu2024rethinking, huang2024localmamba} or construct different graph relationships\cite{behrouz2024graphmambaACM, wang2024graph} to represent the practical spatial distributions.

To promote the sequence representation in pixel-wise correlation, we perform an FRFT on the input and then four distinct scanning paths: including diagonal orientations: top left to bottom right (left-right), top left to bottom right (top-down), and their inverse directions (right-left and down-top) along both the spatial domain and the immediate FRFT domain.
This procedure can be encapsulated as:
\begin{equation}
    \begin{gathered}
    \mathbf{F}_{1} = \mathbf{M}_{3 \times 3}(\mathbf{X}), \mathbf{F}_{Spa}  = \mathbf{M}_{scan}(\mathbf{F}_{1}) \\
    \mathbf{F}_{FRFT}  = \mathbf{M}_{iFRFT}(\mathbf{M}_{scan}(\mathbf{M}_{FRFT}(\mathbf{F}_{1}))) \\
    \mathcal{P}_{spa} = \mathbf{M}_{3 \times 3} (\mathbf{M}_{ctc} (\mathbf{F}_{spa}, \mathbf{F}_{FRFT}) \cdot  \mathbf{M}_{SiLU}(\mathbf{F}_{1}) \\
    \end{gathered}
\end{equation}
where $\mathbf{M}_{3 \times 3}$, $\mathbf{M}_{scan}$, $\mathbf{M}_{iFRFT}$, $\mathbf{M}_{FRFT}$, $\mathbf{M}_{ctc}$, $\mathbf{M}_{SiLU}$ individually denote the operations of $3 \times 3$ convolution, SSM scanning process, iFRFT, FRFT calculation, concatenation, and SiLU function.
Besides, $\mathbf{F}_{1}$, $\mathbf{F}_{Spa}$, and $\mathbf{F}_{FRFT}$ respectively represent the shallow feature, further features in the spatial and FRFT domains. 
Furthermore, the final spatial representation $\mathcal{P}_{spa}$ can be depicted as the fusion of $\mathbf{F}_{Spa}$ and $\mathbf{F}_{FRFT}$.
In this way, $\mathcal{P}_{spa}$ in the FRFT domain serves as a bridge between spatial and Fourier spaces to delineate abundant spatial detail in an intermediate domain.

\begin{figure}[!t]
	\centering
    \includegraphics[width=3.0in]{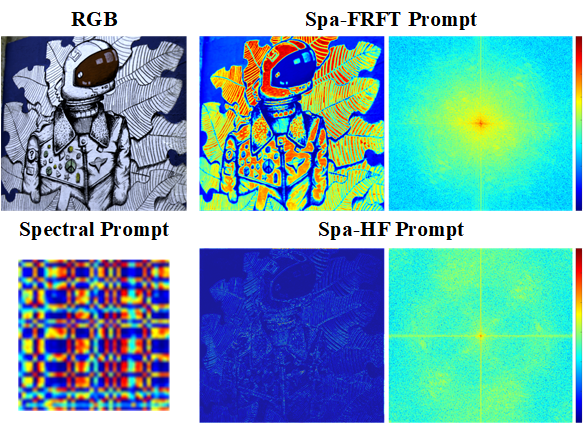}%
    \caption{\textbf{Visulaization of multi-modality features from DRPN.} Specifically, we visualize the frequency components of Spa-FRFT Prompt $\mathcal{P}_{spa}$ and Spa-HF Prompt $\mathcal{P}_{hf}$. It indicates that $\mathcal{P}_{spa}$ focuses on the center, representing the low-frequency global spatial-wise content; while $\mathcal{P}_{hf}$ on the periphery, denoting the high-frequency edge details. Besides, Spectral Prompt $\mathcal{P}_{spe} \in R^{C \times C}$ denotes intrinsic spectral-wise dependency. }
    \label{fig_vis_drpn}
\end{figure}

\textbf{Spa-HF Prompt $\mathcal{P}_{hf}$}: 
As shown in Fig. \ref{fig_vis_drpn}, we found that $\mathcal{P}_{spa}$ favors global spatial content, which brings a certain degree of losing edge details, hiding in the high-frequency components.
Therefore, a compensating branch is designed to delineate $\mathcal{P}_{hf}$.
In detail, average pooling and sequential upsampling operations are leveraged to maintain most of the spatial content $\mathbf{F}_{2}$ from the former feature $\mathbf{F}_{1}$.
Then, the high-frequency details can be peeled off from the 
disparity of $\mathbf{F}_{1}$ and $\mathbf{F}_{2}$.
At last, two $\mathbf{M}_{3 \times 3}$ and a GeLU are utilized again to obtain the final $\mathcal{P}_{hf}$:
\begin{equation}
    \begin{gathered}
    \mathbf{F}_{2} =  \mathbf{M}_{up}(\mathbf{M}_{avg}(\mathbf{F}_{1}))  \\
    \mathcal{P}_{hf} = \mathbf{M}_{3 \times 3} (\mathbf{M}_{GeLU} (\mathbf{M}_{3 \times 3}(\mathbf{F}_{1} - \mathbf{F}_{2}) )) \\
    \end{gathered}
\end{equation}
where $\mathbf{M}_{up}(\cdot)$ and $\mathbf{M}_{avg}(\cdot)$ respectively denote the average pooling and upsampling operations.

\textbf{Spectral Prompt $\mathcal{P}_{spe}$}:
With $\mathbf{X}$, $\mathcal{P}_{spa}$, and $\mathcal{P}_{hf}$, the former Gamma-modeled point spectra $\mathbf{Y}_{g}^{p}$ can be delineated with spatial distribution.
In specific, the average of RGB is concatenated with the $\mathbf{Y}^{p}_{g}$, and then two convolutions are used to model a shallow HSI $\mathbf{Y}^{1} \in R^{B \times H \times W}$:
\begin{equation}
\begin{aligned}
    & \mathbf{X}_{m}  = (\mathbf{X}_{R} + \mathbf{X}_{G} + \mathbf{X}_{B}) / 3 \\
    \mathbf{Y}^{1}  =& \mathbf{M}_{3 \times 3} (\mathbf{M}_{GeLU} (\mathbf{M}_{3 \times 3}(\mathbf{M}_{cct} (\mathbf{Y}^{p}_{g},\mathbf{X}_{m} )) 
\end{aligned}
\end{equation}
where $\mathbf{X}_{m}$ represents the mean of Red 
($\mathbf{X}_{R}$), Green ($\mathbf{X}_{G}$), and Blue ($\mathbf{X}_{B}$) bands of $\mathbf{X}$.

Next, Spa-FRFT Prompt $\mathcal{P}_{spa}$ from the first branch is leveraged to concatenate with $\mathbf{Y}^{1}$ for spatial refinement. Then, Spa-HF Prompt $\mathcal{P}_{hf}$ from the second branch is fed with $\mathbf{Y}^{2}$ to compensate for high-frequency details:
\begin{equation}
    \begin{aligned}
    \mathbf{Y}^{2}  &= \mathbf{M}_{3 \times 3}(\mathbf{M}_{cct} (\mathbf{Y}^{1}, \mathcal{P}_{spa} ))    \\
    \mathbf{Y}^{3}  &= \mathbf{M}_{3 \times 3}(\mathbf{M}_{cct} (\mathbf{Y}^{2}, \mathcal{P}_{hf} ))  \\
    \end{aligned}
\end{equation}
where $\mathbf{Y}^{3}$ represents a preliminary reconstruction of HSI, which is also sent to the sequent DyPro-Mamba for guidance.
Finally, the spectral prompt $\mathcal{P}_{spe} \in R^{C \times C}$ is modeled via a spectral-wise self-attention (SSA)\cite{mst, cai2022mst++}.


\subsubsection{DyPro-Mamba.}
As shown in Fig. \ref{fig_vis_drpn}, features from DRPN embody diverse aspects of available $ \mathbf{X} $ and $ \mathbf{Y}^{p} $, which subsequently undergo elaborate processing in the innovative DyPro-Mamba. 
First, a  $\mathbf{M}_{3 \times 3}(\cdot)$ is used to map $\mathbf{X}$ to a latent feature space.
Then, the output is fused with multi-type representation $\mathbf{Y}^{3}$ from DRPN, which assures dependable guidance from both spatial and spectral dimensions.
Next, three PromptSSMs with corresponding downsample and upsample operations are performed on $\mathbf{Y}^{4}$ to refine the extracted features.
Finally, a $\mathbf{M}_{3 \times 3}(\cdot)$ is leveraged to achieve a final mapping to obtain the reconstructed HSI $\mathbf{Y}^{r}$.
The overall processes can be represented as:
\begin{equation}
\begin{aligned}
    &\mathbf{Y}^{4} = \mathbf{M}_{3 \times 3}(\mathbf{M}_{cct} (\mathbf{M}_{3 \times 3}(\mathbf{X}), \mathbf{Y}^{3})) \\
    &\mathbf{Y}^{5} = \mathbf{M}_{us}(\mathbf{M}_{ps}(\mathbf{M}_{ds}(\mathbf{M}_{ps}(\mathbf{Y}^{4})))) \\
    & \mathbf{Y}^{r} = \mathbf{M}_{3 \times 3}(\mathbf{M}_{ps}((\mathbf{Y}^{5}))) \\
\end{aligned}
\end{equation}
where $\mathbf{M}_{ps}(\cdot)$ and $\mathbf{M}_{ds}(\cdot)$ respectively denote the PromptSSM and downsample.

Specifically, each PromptSSM contains a spectral-prompt SSA, a further spatial SSM, and a Feed-Forward Network (FFN). 
In detail, given an input feature $\mathbf{F}_{in}^{1}  \in R^{H \times W \times C} $, corresponding tokens $\mathbf{F}_{in}^{2}  \in R^{HW \times C}$ are obtained with a reshape operation and a linear layer.
Then, three $\mathbf{M}_{3 \times 3}$ are performed to project 
$\mathbf{F}_{in}^{2}$ into query $\mathbf{Q} \in R^{HW \times \frac{B}{N}}$, key $\mathbf{K} \in R^{HW \times \frac{B}{N}} $, and value $\mathbf{V} \in R^{HW \times \frac{B}{N}}$ elements, where N denotes the number of multi-heads.
With the assistance of $\mathcal{P}_{spe}$ for spectral-wise refinement, the global spectral dependency can be calculated as:
\begin{equation}
\mathbf{F}_{in}^{3} = \mathbf{V} \cdot (\mathcal{P}_{spe} \cdot \mathbf{M}_{sft} (\mathbf{K}^{T} \mathbf{Q}))  
\end{equation}
Next, an SSM is performed to scan $\mathbf{F}_{in}^{3}$ for spatial feature extraction. 
At last, an efficient FFN, consisting of several $\mathbf{M}_{3 \times 3}$ and $\mathbf{M}_{GeLU}$ to form the final output $\mathbf{F}_{out}$.

In summary, this orchestrated approach with its emphasis on multi-type and dynamic prompt generation, constitutes a robust and versatile framework for Pixel-SSR, poised to make significant contributions to the field.


\section{Experiment}
\label{sec:experiment}
This section presents a comprehensive experimental evaluation of our method.
Section \ref{abla_gamma} describes the experimental settings.
In Section \ref{unsupervised comparison}, qualitative and quantitative comparison experiments are performed with state-of-the-art methods in unsupervised mode to explore the improvements with limited label consumption.
Section \ref{universal paradigm} reports reconstruction experiments that assess the effectiveness of Pixel-SSR as a universal paradigm integrating various methods, comparing with unsupervised and image-level supervised modes. 
Finally, Section \ref{section_ablation} presents an ablation study that examines the contributions of the Gamma-modeled strategy to Pixel-SSR’s implementation and DyPro-Mamba to the enhancement of HSI reconstruction performance.

\begin{table*}[htbp]
  \centering
  \renewcommand\arraystretch{1.5}
  \resizebox{\linewidth}{!}{
    \begin{tabular}{cccccccccccccc}
    \toprule[1.5pt]
    \multirow{2}[2]{*}{Methods} & \multicolumn{1}{c}{Venue/} & \multicolumn{4}{c}{NTIRE} & \multicolumn{4}{c}{CAVE}      & \multicolumn{4}{c}{GF-X} \\
          & \multicolumn{1}{c}{Percentage} & RMSE   & PSNR  & SSIM  & SAM   & RMSE   & PSNR & SSIM  & SAM   & RMSE   & PSNR  & SSIM  & SAM \\
    \midrule
    MST++\cite{cai2022mst++} & CVPR22 & 5.03  & 30.42 & 0.969 & 4.50  & 4.20   & 30.48 & 0.827 & 9.56  & 5.48  & 25.88 & 0.525 & 30.60 \\
    Restormer\cite{restormerCVPR2022} & CVPR22 & 5.44  & 28.72 & 0.952 & 5.04  & 4.18  & 29.82 & 0.795 & 11.28 & 6.11  & 24.81 & 0.406 & 35.35 \\
    SAUNET\cite{wang2023saunetACM} & ACM23 & 4.59  & 31.50 & 0.966 & 5.73  & 4.96  & 30.22 & 0.824 & 12.24 & 6.09  & 24.49 & 0.332 & 36.78 \\
    SPECAT\cite{yao2024specatCVPR} & CVPR24 & 4.88  & 30.90 & 0.964 & 4.61  & 5.88  & 27.79 & 0.744 & 14.74 & 5.48  & 25.54 & 0.410 & 31.06 \\
    RtxMamba\cite{bai2024retinexmamba} & Arxiv24 & 5.11  & 29.17 & 0.934 & 5.20  & 6.15  & 27.11 & 0.641 & 14.59 & 4.81  & 26.82 & 0.523 & 28.54 \\
    MFSN\cite{wu2024MSFNTnnls}  & TNNLs24 & 6.63  & 27.92 & \underline{0.973} & 3.23  & 4.59  & 30.45 & 0.847 & 9.38  & 4.45  & 27.49 & 0.626 & 25.92 \\
    UnSSR\cite{leng2025uncertaintyTnnls} & TNNLs25 & \underline{4.22}  & \underline{32.42} & \underline{0.973} & \underline{4.87}  & \underline{3.42}  & \underline{32.58} & \underline{0.946} & \underline{7.66}  &  \underline{3.81}  & \underline{28.75} & \underline{0.688} & \underline{19.39} \\
    \midrule
    \multirow{4}[2]{*}{Ours} & 0.01\% & 1.43  & 37.73 & 0.976 & 2.93  & 1.47  & 39.58 & 0.976 & 4.54  & 1.67  & 35.57 & 0.921 & 6.81 \\
          & 0.10\% & 1.23  & 39.91 & 0.978 & 2.76  & 1.47  & 39.73 & 0.976 & 4.49  & 1.00  & 40.05 & 0.968 & 4.81 \\
          & 1\%   & 1.25  & 40.32 & \textbf{0.981} & 2.69  & 1.43  & 39.84 & 0.973 & 4.53  & 0.47  & 47.03 & 0.991 & 2.71 \\
          & 10\%  & \textbf{1.21}  & \textbf{40.73} & 0.980 & \textbf{2.62}  & \textbf{1.38}  & \textbf{40.15} & \textbf{0.980} & \textbf{4.31}  & \textbf{0.44}  & \textbf{47.52} & \textbf{0.991} & \textbf{2.57} \\
    \bottomrule
    \end{tabular}%
    }
    \caption{\textbf{Testing results on two visual and one remote sensing datasets: NTIRE, CAVE, and GF-X.} The overall best results are $\mathbf{highlighted}$ and the best results among unsupervised methods are \underline{underlined}.
  The percentages indicate the proportion of point spectra used as labels relative to an entire HSI. 
  During our and compared training, $128 \times 128$ patch is used as a minimum input, and ``0.01\%" means that  ``$ \lfloor 128 \times 128 \times 0.01\% \rfloor = 1 $" point spectra in a patch is used as the label. 
  For clearer comparisons, the actual RMSE values are obtained by multiplying the values in the table by $10^{-2}$, for example, the RMSE of MST++ in NTIRE is $ 5.03 \times 10^{-2} = 0.0503$.
  }
  \label{com_results}%
\end{table*}%

\subsection{Settings}
\label{settings}

\textbf{Hyperspectral Datasets.}
To evaluate our method, two visual and one practical scenes are adopted, denoted as NTIRE, CAVE, and GF-X. Considering the pixel-level task, we also demonstrate the total pixel numbers.
\begin{itemize}
\item  {
$8.39 \times 10^6$ pixels in CAVE: 32 indoor HSIs with a spatial resolution of $512 \times 512$, ranging from 400–700 nm. 
}
\item  {
$1.18 \times 10^8$ pixels in NTIRE: 450 training and 30 validation images, individually featuring $482 \times 512$, ranging from 400–700 nm. The images depict a diverse range of indoor and outdoor scenes and were captured using the Specim IQ mobile hyperspectral camera.
}
\item  {
$1.72 \times 10^7$ pixels in GF-X: a remote sensing scene from GF-X satellite, encompassing $ 1948 \times 4112$ and  $ 1948 \times 4709$ HSI-MSI pairs, covering 400-1000nm.
}
\end{itemize}

\noindent{\textbf{Implementation Details.}}
In the training procedure, the band number of features in DyPro-Mamba ($B$) is empirically set to 40. 
According to the image size, $128 \times 128$ patches are cropped. 
Take $128 \times 128$ as a minimum input, the minimum percentage is set as 0.01\%, which means ``$ \lfloor 128 \times 128 \times 0.01\% \rfloor = 1 $" point spectra is selected as label rather than the whole $128 \times 128$ patch.
The parameter optimization algorithm chooses Adam with $\beta_{1}=0.9, \beta_{2}=0.99$ and $\epsilon=10^{-8}$. The learning rate is initialized as 0.0002 and the decay policy is set as a power=1.5. 
Our paradigm has been implemented on the Pytorch framework and 200 epochs have been trained on one NVIDIA 3090Ti GPU.

\noindent{\textbf{Evaluation Criteria.}}
To objectively evaluate the  performance, we utilize four metrics: root mean square error (RMSE), peak signal-to-noise ratio (PSNR), structural similarity (SSIM), and spectral angle mapper (SAM). Typically, a lower RMSE or SAM value, coupled with a higher PSNR or SSIM value, signifies better performance.

\subsection{Comparison with SOTA Methods}
\label{unsupervised comparison}

\noindent{\textbf{Quantitative Comparison.}}
We compare our method with state-of-the-art reconstruction methods: MST++ \cite{cai2022mst++}, Restormer \cite{restormerCVPR2022}, SAUNET \cite{wang2023saunetACM}, SPECAT \cite{yao2024specatCVPR}, RtxMamba \cite{bai2024retinexmamba}, MSFN \cite{wu2024MSFNTnnls}, and UnSSR \cite{leng2025uncertaintyTnnls}.
Table \ref{com_results} reports the numerical results on all datasets. 
As can be seen, Ours with 0.01\% label consumption than an entire HSI, achieves outstanding improvements of 66.11\%, 57.02\%, and 56.17\% over the best-unsupervised methods in RMSE, 5.31dB, 6.70dB, and 6.82dB in PSNR on NTIRE, CAVE, and GF-X, respectively.
When the label ratio reaches 10\%, the performance in RMSE is improved to 71.33\%, 59.65\%, and 88.45\%, that in PSNR is improved to 8.31dB, 7.56dB, and 18.77dB.
We can reasonably deduce that our Pixel-SSR acts as a powerful and efficient strategy for HSI reconstruction.

\noindent{\textbf{Qualitative Comparison.}}
As shown in Fig \ref{vis_all}, we performed RMSE heat maps of all datasets, showing higher accuracy with darker colors. 
Additionally, spectral curves were plotted in a random pixel, highlighted as a red circle in the RGB image, with corresponding spectral profiles displayed for comparison. 
Ours exhibits the darkest error maps and the most similar with the groundtruth, signifying its superiority whether from a holistic or detailed aspect.

\begin{table}[!t]
  \centering
  \renewcommand\arraystretch{1.5}
    \resizebox{\linewidth}{!}{
    \begin{tabular}{lc|lc}
    \toprule[1.5pt]
    Methods & PSNR  & Methods & PSNR \\
    \hline
    MST++\cite{cai2022mst++} & 30.42 & RtxMamba\cite{bai2024retinexmamba} & 29.17 \\
    \rowcolor[rgb]{ .851,  .851,  .851} MST++ + \textbf{Ours} & 38.41 & RtxMamba + \textbf{Ours} & 38.07 \\
    Supervised MST++ & 39.90 & Supervised RtxMamba & 41.62 \\
    \hline
    Restormer\cite{restormerCVPR2022} & 28.72 & MFSN\cite{wu2024MSFNTnnls}  & 27.92 \\
    \rowcolor[rgb]{ .851,  .851,  .851} Restormer + \textbf{Ours} & 37.92  & MFSN + \textbf{Ours} & 38.92 \\
    Supervised Restormer & 40.18 & Supervised MFSN & \textbf{42.13} \\
    \hline
    SAUNet\cite{wang2023saunetACM} & 31.50 & UnSSR\cite{leng2025uncertaintyTnnls} & \textbf{32.42} \\
    \rowcolor[rgb]{ .851,  .851,  .851} SAUNet + \textbf{Ours} &  36.58     & UnSSR + \textbf{Ours} & 39.02 \\
    Supervised SAUNet & 37.49 & Supervised UnSSR & 40.17 \\
    \hline
    SPECAT\cite{yao2024specatCVPR} & 30.90 &  & \\
    \cellcolor[rgb]{ .851,  .851,  .851}{SPECAT+ \textbf{Ours} } & \cellcolor[rgb]{ .851,  .851,  .851}{37.69} & \cellcolor[rgb]{ .851,  .851,  .851}{Our DyPro-Mamba}      & \cellcolor[rgb]{ .851,  .851,  .851}{\textbf{39.58}} \\
    Supervised SPECAT & 39.24 &       &  \\
    \bottomrule[1.5pt]
    \end{tabular}%
    }
  \caption{\textbf{Effectiveness as a universal Pixel-SSR paradigm.} All baselines are based on the unsupervised SSR, ``+ \textbf{Ours}" denotes training on our Pixel-SSR with 0.01\% label percentage, and Supervised denotes training on the image-level supervision. The best results of three modes, including 32.42dB(UnSSR), 39.58dB(Our DyPro-Mamba), and 42.13dB(MFSN), are individually \textbf{highlighted}. Our Pixel-SSR improves the results substantially over the unsupervised mode, achieves the best results in the same Pixel-SSR paradigm(Gray), and obtains comparable results with supervised mode with significantly efficient label consumption.}
  \label{com_paradigm}%
\end{table}%

\subsection{Effectiveness as a universal paradigm.}
\label{universal paradigm}
To serve as a universal paradigm, we replaced the dotted box in Fig. \ref{framework} with other leading methods, sharing the same DRPN and loss constraints as our Pixel-SSR.
As shown in Table \ref{com_paradigm} and Fig. \ref{fig1}, all methods have been conducted in Our Pixel-SSR with 0.01\% label consumption to compare with those on unsupervised (without label) and supervised (with total entire images as the label) modes. 
We can infer that 1) our Pixel-SSR improves the results substantially over the unsupervised mode, 2) achieves the best results than other methods in the same Pixel-SSR paradigm(Gray), and 3) obtains comparable results with image-level supervised mode with efficient label consumption.

\subsection{Ablation Studies}
\label{section_ablation}
We conducted ablations on the CAVE dataset. More detailed results are provided in the supplementary.

\begin{table}[!t]
  \centering
    \begin{tabular}{ccccc}
    \toprule[1.5pt]
    PSNR($\uparrow$)  & 0.01\% & 0.10\% & 1\%   & 10\%  \\
    \hline
    Gaussian & 36.83 & 37.97 & 39.26 & 40.09 \\
    Gamma(Ours) & \textbf{39.58} & \textbf{39.73}  & \textbf{40.15} & \textbf{40.21} \\
    \bottomrule[1.5pt]
    \end{tabular}%
    \caption{Ablation study on different label percentages to investigate the influence of Gaussian- or Gamma-modeled point spectra.}
  \label{abla_gamma}%
  
\end{table}%

    
    
    

\begin{figure}[htbp]
	\centering
	\includegraphics[width=3.3in]{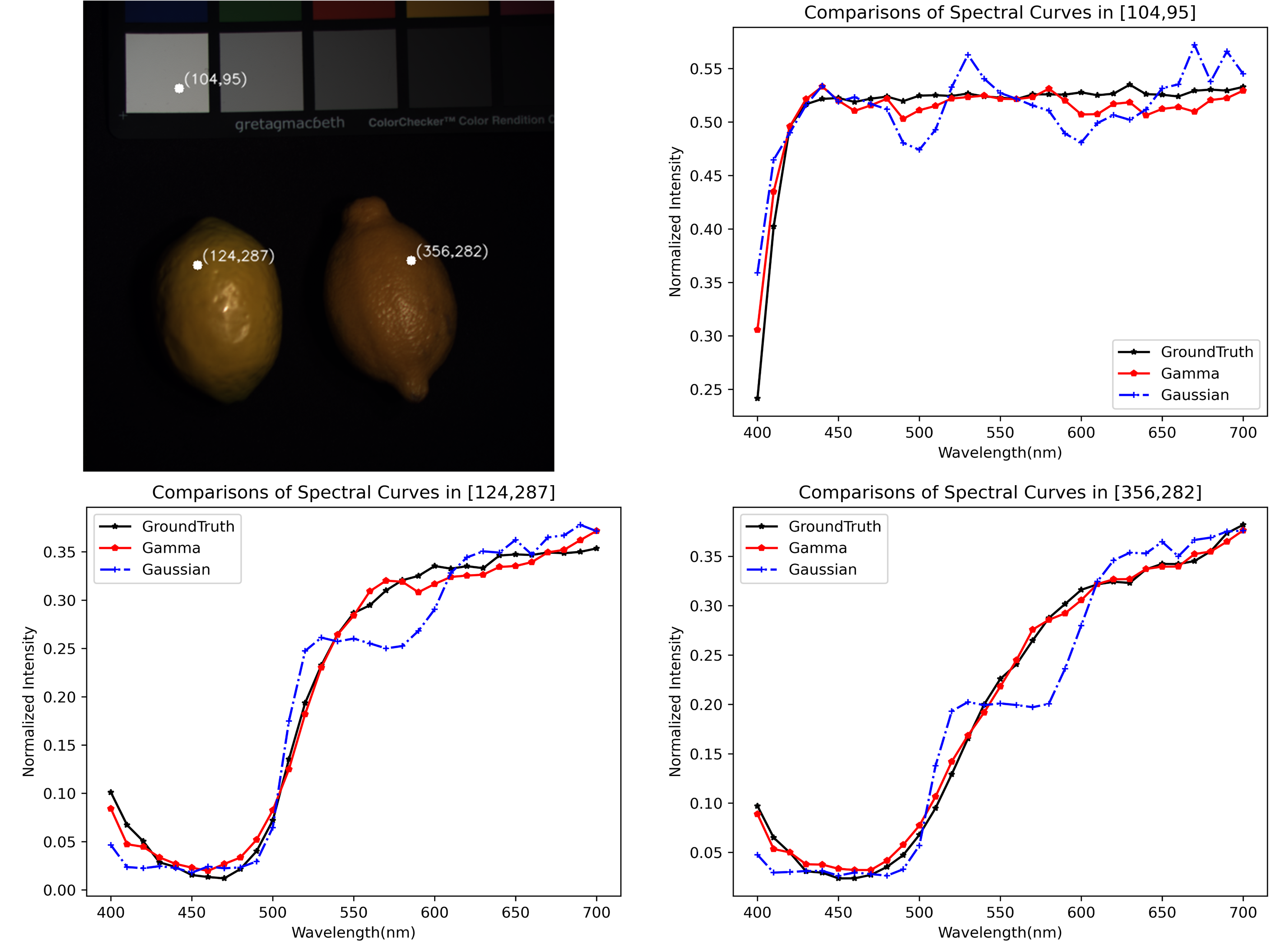}%
    \caption{\textbf{Comparisons of reconstructed spectral curves among Gamma- and Guassian-modeled strategies.} Three representative points are selected to analyze the accuracy in background, fake lemons, and real lemons, indicating that Gamma-modeled strategy achieves more precise and high-fidelity results, especially when distinguishing slight gaps between real and fake objects.}
    \label{gamma_verify}
\end{figure}

\noindent{\textbf{Investigation of Gamma-modeled Point Spectra.}}
As listed in Table \ref{abla_gamma}, modeling point spectra using a Gaussian distribution resulted in different degrees of performance decline for different ratios. 
Besides, we visualized the reconstructed spectral curves of Gamma-modeled and Gaussian-modeled results in Fig \ref{gamma_verify}, showing Gamma achieved more similar and precise curves with the groudtruth, especially high-fidelity results in distinguish slight gaps between real and fake objects.
We hypothesize that this advantage arises from the Gamma distribution's capacity to effectively model signals with intensity variations, such as spectral bands with skewed distributions. 

\begin{table}[htbp]
  \centering
  \renewcommand\arraystretch{1.5}
  \resizebox{\linewidth}{!}{
    \begin{tabular}{cccccccc}
    \toprule[1.5pt]
    \multicolumn{3}{c}{DRPN} & \multirow{2}[0]{*}{PromptSSM} & \multirow{2}[0]{*}{Params(M)} & \multirow{2}[0]{*}{Flops(G)} & \multirow{2}[0]{*}{PSNR} & \multirow{2}[0]{*}{SAM} \\
    \cline{1-3}
   $ \mathcal{P}_{spa}$  & $\mathcal{P}_{hf}$    & $\mathcal{P}_{spe}$   &       &       &       &       &  \\
    \hline
    & \checkmark      & \checkmark     & \checkmark     & 0.61  & 4.6   & 39.04 & 4.84  \\
    \checkmark     &      & \checkmark      & \checkmark     & 0.61  & 4.63  & 39.71 & 4.45 \\
    \checkmark &       \checkmark &   & \checkmark  & 0.57  & 4.24  & 39.72 & 4.50 \\
    &      &      & \checkmark        & 0.56  & 4.18  & 39.74 & 4.50  \\
    \checkmark     & \checkmark     & \checkmark     &       & 0.56  & 3.92  & 39.99 & 4.52  \\
    \checkmark & \checkmark & \checkmark     & \checkmark     & 0.61  & 4.65  & \textbf{40.15} & \textbf{4.31} \\
    \bottomrule[1.5pt]
    \end{tabular}%
    }
    \caption{Ablation study to investigate the components of our proposed DyPro-Mamba, depicted in the former four columns. }
  \label{Ablation}%
\end{table}%

\noindent{\textbf{Investigation of $\mathcal{P}_{spa}$ branch.}}
When we delete the spatial branch $\mathcal{P}_{spa}$, the model performance dropped 1.1044dB on PSNR and 12.43\% on SAM. The results indicate that the intermediate space contains detailed non-stationary content, which depicts abundant global spatial-wise distributions that are vital to the reconstruction performance.

\begin{figure*}[!t]
	\centering
	\includegraphics[width=0.98\textwidth]{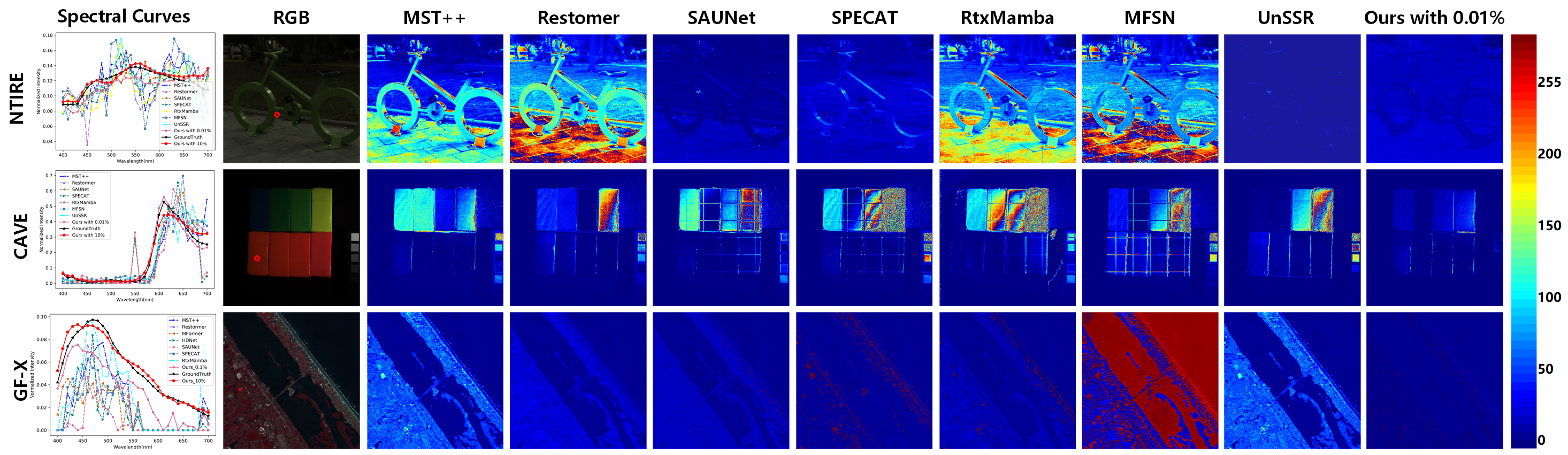}%
	\caption{ \textbf{Qualitative results on NTIRE, CAVE, and GF-X datasets.} Comparisons of the RMSE error maps of the reconstructed HSI from MST++/Restormer/SAUNet/SPECAT/RtxMamba/MFSN/UnSSR/Ours with 0.01\%. Besides, the spectral curves in a randomly selected pixel (marked as the red circle in RGB) are drawn in the left. Ours with 0.01\% label consistently exhibits the darkest error maps and achieves the most similar spectral curves with the groundtruth, signifying its superiority whether from a holistic or detailed aspect. 
    }
    \label{vis_all}
\end{figure*}

\noindent{\textbf{Investigation of $\mathcal{P}_{hf}$ branch.}}
When we remove the high-frequency $\mathcal{P}_{hf}$ branch, the performance is reduced by 0.4369dB on PSNR and 3.43\% on SAM, showing the importance of edge details hidden in high-frequency space.

\noindent{\textbf{Investigation of $\mathcal{P}_{spe}$ branch.}}
When we remove the spectral-wise dependancy $\mathcal{P}_{hf}$ branch, leaving $\mathcal{P}_{spa}$ and $\mathcal{P}_{hf}$. 
The performance decreased by 0.4226dB on PSNR and 3.56\% on SAM without the guidance of 
pre-extracted spectral guidance, indicating that the extracted spectral-wise dependency is vital for the Pixel-SSR task.

\noindent{\textbf{Investigation of DRPN.}}
When deleting DRPN, PSNR and SAM respectively dropped by 0.4047dB and 4.46\%, which demonstrates that our DRPN with 0.05M parameters and 0.47G flops, improves performance via effectively mining available information from RGB and point spectra.

\noindent{\textbf{Investigation of PromptSSM.}}
When removing PromptSSM, the accuracy decreased by 0.4005dB on PSNR and 7.48\% on SAM, indicating the necessity of the guidance of spectral-wise dependency.

\begin{table}[h]  
	\centering
    \label{FRFT}
	\begin{tabular}{ccccc} 
        \hline
        \specialrule{0em}{0.5pt}{0.5pt}
         Setting  &  PSNR($\uparrow$)  & SAM($\downarrow$) \\
		\hline 	 
	   $ \mathcal{P}_{spa}$ with FFT  &  39.64 & 4.61\\
	$ \mathcal{P}_{spa}$ with FRFT(Ours)  & \textbf{40.15}  & \textbf{4.31}\\ 
		\bottomrule 
	\end{tabular}	
\caption{Ablation study on the influence of FRFT in $\mathcal{P}_{spa}$.}
\label{FRFT}
\end{table}

\noindent{\textbf{Investigation of FRFT in $\mathcal{P}_{spa}$.}}
As listed in Table \ref{FRFT}, the results decreased by 0.51dB on PSNR and 6.51\% on SAM, verifying that FRFT performs better for Pixel-SSR than FFT due to its ability to process non-stationary signals flexibly rotate between time and frequency domains.


\section{Conclusion}
\label{sec:conclusion}
In this work, we explore the Pixel-SSR task to achieve a robust balance between reconstruction accuracy and label efficiency. 
First, we introduce a Gamma-modeled strategy to generate point spectra, enhancing adaptability and deployment in new scenes where point spectra are unavailable. 
Then, our DyPro-Mamba gradually delineates complementary features from RGB and point spectra and then fuses these multi-type features to endow the reconstructed HSI with refined spatial and spectral details. 
In spite of outstanding performance in HSI reconstruction with exhaustive horizontal and vertical experiments, our DyPro-Mamba also provides a spectral enhancement strategy for RGB-based downstream tasks, such as classification, segmentation, target detection, etc.

{
    \small
    \bibliographystyle{ieeenat_fullname}
    \bibliography{main}
}

\clearpage
\setcounter{page}{1}
\maketitlesupplementary

The supplementary is organized as follows:
\begin{itemize}
    \item {Section \ref{evaluation}: The mathematical formulas of evaluation criteria.}
    \item {Section \ref{section_loss}: Formulation and Investigation of Loss function.}
    \item {Section \ref{sup_com}: More quantitative results to verify model stability and efficiency.}
    \item {Section \ref{paradigm}: More quantitative results to verify the effectiveness of our Pixel-SSR as a universal paradigm.}
    \item {Section \ref{sup_abla}: More ablation results on Gamma-modeled strategy and our proposed DyPro-Mamba.}
    \item {Section \ref{sup_vis}: More qualitative details of error maps and feature visualization in DRPN.}
\end{itemize}

\section{The mathematical formulas of evaluation criteria}
\label{evaluation}
Four adopted evaluation metrics in manuscripts are root mean square error (RMSE), peak signal-to-noise ratio (PSNR), structural similarity (SSIM), and the Spectral Angle Mapper (SAM). The related formulas are as follows:
\begin{align}
    \operatorname{RMSE}&=\sqrt{\frac{1}{N} \sum_{p=1}^{N}\left(\mathbf{Y}_{ori}[p]-\mathbf{Y}_{res}[p]\right)^{2}} \\
    \operatorname{PSNR}&= - \frac{20}{N} \sum_{p=1}^{N} \log \left( \left| \mathbf{Y}_{ori}[p]-\mathbf{Y}_{res}[p] \right| \right) \\ 
    \operatorname{SSIM}&= \frac{1}{D} \sum_{d=1}^{D}\operatorname{SSIM} \left(\mathbf{Y}_{ori}[d], \mathbf{Y}_{res}[d]\right) \\
    \mathrm{SAM} & = \frac{1}{B} \sum_{bi=1}^{B}\left(\frac{180}{\pi} \arccos \frac{\left\langle \mathbf{Y}_{ori}[bi], \mathbf{Y}_{res}[bi] \right\rangle}{\left\|\mathbf{Y}_{ori}[bi] \right\|_2\left\|\mathbf{Y}_{res}[bi]\right\|_2}\right)
\end{align}

where $\left(\mathbf{Y}_{ori}[p],\mathbf{Y}_{res}[p]\right)$ and $\left(\mathbf{Y}_{ori}[d],\mathbf{Y}_{res}[d]\right)$ denote the p-th pixel and d-th band of the groundtruth and reconstructed HSI respectively. $N$ and $D$ are the sums of pixels and bands of the HSI cube. 
Concretely, RMSE signified the disparity in pixel dimension between the groundtruth and recovered HSI, PSNR reveals the image quality of recovered HSIs, SSIM indicates the spatial structure similarity between two images, while SAM demonstrates the spectral similarity of all pixels between two images. 
In general, smaller RMSE and SAM, or larger PSNR and SSIM demonstrate more outstanding performance.

\section{Formulation and Investigation of Loss function.}
\label{section_loss}
To drive DyPro-Mamba, four customized losses are adopted to model the Pixel-SSR task, including a projected loss $\mathcal{L}_{rep}$, a pixel spectra-based loss $\mathcal{L}_{pHSI}$, an SSIM-based loss $\mathcal{L}_{SSIM}$ \cite{li2023mformer}, and a Pixel2Image loss $\mathcal{L}_{P2I}$.
\begin{equation}
\begin{gathered}
\mathcal{L}_{rep}  = \left\| \mathbf{X} - \mathbf{S} \mathbf{Y}^{r} \right\|_1, \mathcal{L}_{pHSI}  = \left\| \mathbf{Y}^{r}  - \mathbf{Y}^{p}  \right\|_1 \\
 \mathcal{L}_{SSIM} = 1 - SSIM (\mathbf{Y}^{r}(i), \mathbf{Y}^{r}(i+1)), \\ \mathcal{L}_{P2I}  = \left\| \mathbf{X}  - \mathbf{M}_{H2R}(\mathbf{Y}^{3})  \right\|_1 \\
\end{gathered}
\end{equation}
where $i$ denotes the band index and $\mathbf{M}_{H2R}(\cdot)$ represents our mapping module that constrains $\mathbf{Y}^{3}$ derived from the DRPN using the input RGB, which consists of two $3 \times 3$ convolutions and a GeLU function. 
Thus, $\mathcal{L}_{P2I}$ approximate the process from random Gamma-based point spectra to the real HSI in the spatial dimension.
Finally, the overall loss function can be expressed as follows:
\begin{equation}
\mathcal{L}_{overall} = \mathcal{L}_{pHSI} + \beta_{1} \mathcal{L}_{rep} + \beta_{2} \mathcal{L}_{SSIM} + \beta_{3} \mathcal{L}_{P2I}
\end{equation}
where $ \{ \beta_{1}, \beta_{2}, \beta_{3} \}$ are respectively set as 1, 0.3, and 0.3 in our practical settings empirically.

In Table \ref{abla_loss}, we conduct a granular analysis of the impact of each loss function, revealing that the exclusion of any individual loss function results in a decrement in performance metrics, thereby underscoring their indispensable contributions.
\begin{table}[htbp]
  \centering
  \renewcommand\arraystretch{1.5}
  \resizebox{\linewidth}{!}{
    \begin{tabular}{ccccc}
    \hline
    Loss Function & RMSE $(\downarrow)$ & PSNR $(\uparrow)$ & SSIM $(\uparrow)$ & SAM $(\downarrow)$ \\
    \hline
    w/o $\mathcal{L}_{P2I}$  & 0.0145 & 39.6899 & 0.9797 & 4.4689 \\
    w/o $\mathcal{L}_{pHSI}$ & 0.0363 & 31.6965 & 0.8880 & 7.7378 \\
    w/o $\mathcal{L}_{rep}$  & 0.0148 & 39.1505 & 0.9663 & 4.7906 \\
    w/o $\mathcal{L}_{SSIM}$ & 0.0159 & 38.5161 & 0.9697 & 5.0349 \\
    \textbf{Ours} & \textbf{0.0138} & \textbf{40.1464} & \textbf{0.9802} & \textbf{4.3060} \\
    \hline
    \end{tabular}%
    }
  \caption{Ablation study of the influence of loss functions.}
  \label{abla_loss}%
\end{table}%

\begin{table*}[!h]
  \centering
  \renewcommand\arraystretch{1.5}
  \resizebox{\linewidth}{!}{
    \begin{tabular}{ccccc}
    \toprule
    Ratio & RMSE$(\downarrow)$   & PSNR$(\uparrow)$  & SSIM$(\uparrow)$  & SAM$(\downarrow)$ \\
    \hline
    Ours\_0.01\% & $0.0143 \pm 1.169\times 10^{-12}$ & $37.7298 \pm 2.15\times 10^{-7}$ & $0.9759  \pm 1.93 \times 10^{-11}$ & $2.9287  \pm 4.1 \times 10^{-7}$ \\
    Ours\_0.1\% & $0.0123 \pm 2 \times 10^{-16}$ & $39.9135  \pm 2.35 \times 10^{-10}$ & $0.9775  \pm 4 \times 10^{-15}$ & $2.7641  \pm 3.55 \times 10^{-11}$ \\
    Ours\_1\% & $0.0125  \pm 2.2 \times 10^{-14}$ & $40.3217 \pm 82.07 \times 10^{-9}$ & $0.9807 \pm 2.81 \times 10^{-14}$ & $2.6949 \pm 1.45 \times 10^{-10}$ \\
    Ours\_10\% & $0.0121  \pm 1.65 \times 10^{-13}$ & $40.7316  \pm 2.23 \times 10^{-7}$ & $0.9797  \pm 6.23 \times 10^{-13}$ & $2.6197  \pm 8.52 \times 10^{-9}$ \\
    \bottomrule
    \end{tabular}%
    }
    \caption{\textbf{Stability Valuation on NTIRE datasets.} All results are tested ten times, where the variances of all criteria are below \textbf{$10^{-7}$} orders of magnitude. The results indicate that our method addresses the challenge of synthesizing point spectra for new scenes, effectively bridging this gap without inducing significant perturbations to the network's stability. }
  \label{sup_junzhi_ntire}%
\end{table*}%

\begin{table*}[!h]
  \centering
  \renewcommand\arraystretch{1.5}
  \resizebox{\linewidth}{!}{
    \begin{tabular}{ccccc}
    \toprule
    Ratio & RMSE$(\downarrow)$   & PSNR$(\uparrow)$  & SSIM$(\uparrow)$  & SAM$(\downarrow)$ \\
    \hline
    Ours\_0.01\% & $0.0147 \pm 6 \times 10^{-16}$ & $39.5787 \pm 7.79 \times 10^{-10}$ & $0.9759 \pm 6.14 \times 10^{-14}$ & $4.5436 \pm 1.68 \times 10^{-13}$ \\
    Ours\_0.1\% &$ 0.0147 \pm 1 \times 10^{-16}$ & $39.7281 \pm 4.90 \times 10^{-11}$ & $0.9758 \pm 2.6 \times 10^{-15}$ & $4.4857 \pm 1.04 \times 10^{-10}$ \\
    Ours\_1\% & $0.0143 \pm 1 \times 10^{-16}$ & $39.8368 \pm 5.63 \times 10^{-11}$ & $0.9725 \pm 6.01 \times 10^{-14}$ & $4.5345 \pm 1.07 \times 10^{-10}$ \\
    Ours\_10\% & $0.0138 \pm 8 \times 10^{-16}$ & $40.1464 \pm 1.39 \times 10^{-10}$ & $0.9802 \pm 2 \times 10^{-15}$ & $4.306  \pm 4.93 \times 10^{-11} $\\
    \bottomrule
    \end{tabular}%
    }
    \caption{\textbf{Stability Valuation on CAVE datasets.} All results are tested ten times, where the variances of all criteria are below \textbf{$10^{-10}$} orders of magnitude. The results indicate that our method addresses the challenge of synthesizing point spectra for new scenes, effectively bridging this gap without inducing significant perturbations to the network's stability. }
  \label{sup_junzhi_cave}%
\end{table*}%

\begin{table*}[!h]
  \centering
  \renewcommand\arraystretch{1.5}
  \resizebox{\linewidth}{!}{
    \begin{tabular}{ccccc}
    \toprule
    Ratio & RMSE$(\downarrow)$   & PSNR$(\uparrow)$  & SSIM$(\uparrow)$  & SAM$(\downarrow)$ \\
    \hline
    Ours\_0.01\% & $0.0167 \pm 4 \times 10^{-15}$ & $37.5713 \pm 1.05 \times 10^{-9}$ & $0.9213 \pm 1.68 \times 10^{-11}$ & $10.7524 \pm 5.43 \times 10^{-8} $\\
    Ours\_0.1\% & $0.010 \pm 4 \times 10^{-15}$ & $40.0503 \pm 4.12 \times 10^{-9}$ & $0.9676 \pm 2.52 \times 10^{-12}$ & $4.8097 \pm 1.19 \times 10^{-8}$ \\
    Ours\_1\% & $0.0047 \pm 1 \times 10^{-15}$ & $47.0272 \pm 5.88 \times 10^{-8}$ & $0.9908 \pm 1.62 \times 10^{-12}$ & $2.7122 \pm 4.90 \times 10^{-9}$ \\
    Ours\_10\% & $0.0044 \pm 2.1 \times 10^{-15}$ & $47.5221 \pm 5.24 \times 10^{-10}$ & $0.9910 \pm 4.89 \times 10^{-15}$ & $2.2689 \pm 7.65 \times 10^{-11} $\\
    \bottomrule
    \end{tabular}%
    }
    \caption{\textbf{Stability Valuation on GF-X datasets.} All results are tested ten times, where the variances of all criteria are below \textbf{$10^{-8}$} orders of magnitude. The results indicate that our method addresses the challenge of synthesizing point spectra for new scenes, effectively bridging this gap without inducing significant perturbations to the network's stability.. }
  \label{sup_junzhi_gf}%
\end{table*}%

\begin{table*}[!h]
  \centering
    \begin{tabular}{ccccccccc}
    \toprule
    $H \times W = 128 \times 128$ & MST++ & Restormer & SAUNET & SPECAT & RtxMamba & MFSN  & UnSSR & Ours \\
    \midrule
    Params(M) & 1.62  & 15.09 & 4.41  & 0.39  & 3.6   & 2.47  & 1.79  & 0.61 \\
    Flops(G) & 5.83  & 21.4  & 11.12 & 3.49  & 8.997 & 32.67 & 22.12 & 4.65 \\
    Testing Time(ms) & 24.1  & 30.26 & 36.62 & 16.51 & 20.6  & 68.24 & 22.14 & 35.13 \\
    \bottomrule
    \end{tabular}%
    \caption{\textbf{Comparisons of parameters, flops, and testing time among various comparable methods}. Ours achieves the second-lowest Params and Flops, and an acceptable testing time. }
  \label{sup_params}%
\end{table*}%

\begin{table*}[htbp]
  \centering
  \resizebox{\linewidth}{!}{
    \begin{tabular}{cccccccccc}
    \toprule[1.5pt]
    \multirow{2}[4]{*}{Methods}  & \multicolumn{4}{c}{Gamma}     & & \multicolumn{4}{c}{Gaussian} \\
    \cline{2-5}  \cline{7-10} 
     & RMSE $(\downarrow)$  & PSNR $(\uparrow)$  & SSIM $(\uparrow)$  & SAM $(\downarrow)$   & & RMSE $(\downarrow)$  & PSNR $(\uparrow)$  & SSIM $(\uparrow)$  & SAM $(\downarrow)$ \\
    \midrule
    Ours(0.01\%)   & \textbf{0.0147} & \textbf{39.5787} & \textbf{0.9759} & \textbf{4.5436} && 0.0212 & 36.8268 & 0.9369 & 6.4337 \\
    Ours(0.1\%)   & \textbf{0.0147} & \textbf{39.7281} & \textbf{0.9758} & \textbf{4.4857} && 0.0180 & 37.9670 & 0.9637 & 6.5882 \\
    Ours(1\%)   & \textbf{0.0143} & \textbf{39.8368} & 0.9725 & \textbf{4.5345} && 0.0159 & 39.2558 & \textbf{0.9778} & 4.6314 \\
    Ours(10\%)   & 0.0138 & \textbf{40.1464} & \textbf{0.9802} & \textbf{4.3060} && \textbf{0.0136} & 40.0876 & 0.9802 & 4.3611 \\
    \bottomrule[1.5pt]
    \end{tabular}%
    }
    \caption{Ablation study to investigate the influence of Gauss-modeled or Gamma-modeled point spectra. For horizontal comparisons, the best results in the same ratio are \textbf{highlighted}.}
  \label{sup_abla_gamma}%
\end{table*}%

\begin{table*}[!h]
  \centering
  \renewcommand\arraystretch{1.5}
  \resizebox{\linewidth}{!}{
    \begin{tabular}{cccccccccc}
    \toprule[1.5pt]
    \multicolumn{3}{c}{DRPN} & \multirow{2}[0]{*}{PromptSSM} & \multirow{2}[0]{*}{Params(M)} & \multirow{2}[0]{*}{Flops(G)} & \multirow{2}[0]{*}{RMSE$(\downarrow)$} & \multirow{2}[0]{*}{PSNR$(\uparrow)$} & \multirow{2}[0]{*}{SSIM$(\uparrow)$} & \multirow{2}[0]{*}{SAM$(\downarrow)$} \\
    \cline{1-3}
   $ \mathcal{P}_{spa}$  & $\mathcal{P}_{hf}$    & $\mathcal{P}_{spe}$    &       &       &       &       &       &       &  \\
   \hline
    & \checkmark      & \checkmark     & \checkmark     & 0.61  & 4.6   & 0.0128 & 39.6916 & 0.9785 & 2.8309 \\
    \checkmark     &      & \checkmark      & \checkmark      & 0.61  & 4.63  & 0.0138 & 38.8232 & 0.9735 & 3.1422 \\
    \checkmark &       \checkmark &   & \checkmark     & 0.57  & 4.24  & 0.0127 & 40.1137 & 0.9801 & 2.8631 \\
    &      &      & \checkmark      & 0.56  & 4.18  & 0.0140 & 38.7094 & 0.9744 & 3.0590 \\
    \checkmark     & \checkmark     & \checkmark     &     & 0.57  & 3.92  & 0.0142 & 38.7417 & 0.9698 & 3.2014 \\
    \checkmark & \checkmark & \checkmark     & \checkmark     & 0.61  & 4.65  & \textbf{0.0121} & \textbf{40.7316} & \textbf{0.9797} & \textbf{2.6197} \\
    \bottomrule[1.5pt]
    \end{tabular}%
    }
    \caption{Ablation study on NTIRE to investigate the components of our proposed DyPro-Mamba, depicted in the former four columns.}
  \label{sup_ala_ntire}%
\end{table*}%

\begin{table*}[!t]
  \centering
  \renewcommand\arraystretch{1.5}
  \resizebox{\linewidth}{!}{
    \begin{tabular}{ccccc|ccccc|ccccc}
    \toprule[1.5pt]
    Unsupervised & RMSE  & PSNR  & SSIM  & SAM   & Pixel-SSR & RMSE  & PSNR  & SSIM  & SAM   & Supervised & RMSE  & PSNR  & SSIM  & SAM \\
    \hline 
    MST++ & 4.2   & 30.48 & 0.827 & 9.56  & MST++ + \textbf{Ours} & 1.62  & 38.41 & 0.906 & 5.27  & MST++ & 1.5   & 39.90 & 0.990 & 3.85 \\
    Restormer & 4.18  & 29.82 & 0.795 & 11.28 & Restormer + \textbf{Ours} & 1.62  & 37.92 & 0.877 & 5.36  & Restormer & 1.47  & 40.18 & \textbf{0.991} & \textbf{3.59} \\
    SAUNet & 4.96  & 30.22 & 0.824 & 12.24 & SAUNet + \textbf{Ours} & 1.87  & 36.58 & 0.867 & 6.38  & SAUNet & 1.77  & 37.49 & 0.979 & 5.94 \\
    SPECAT & 5.88  & 27.79 & 0.744 & 14.74 & SPECAT + \textbf{Ours} & 1.65  & 37.69 & 0.883 & 5.49  & SPECAT & 1.53  & 39.24 & 0.984 & 3.94 \\
    RtxMamba & 6.15  & 27.11 & 0.641 & 14.59 & RtxMamba + \textbf{Ours} & 1.68  & 38.07 & 0.921 & 5.63  & RtxMamba & 1.48  & 41.62 & 0.985 & 4.20 \\
    MFSN  & 4.59  & 30.45 & 0.847 & 9.38  & MFSN + \textbf{Ours} & 1.53  & 38.92 & 0.930 & 4.85  & MFSN  & \textbf{1.39}  & \textbf{42.13} & 0.986 & 3.68 \\
    UnSSR & \textbf{3.42}  & \textbf{32.58} & \textbf{0.946} & \textbf{7.66}  & UnSSR + \textbf{Ours} & 1.64  & 39.02 & 0.945 & 5.03  & UnSSR & 1.51  & 40.17 & 0.984 & 4.45 \\
    \hline
    - & -  & - & - & - & \textbf{Our DyPro-Mamba} & \textbf{1.47} & \textbf{39.58} & \textbf{0.976} & \textbf{4.54} & - & - & - & - & - \\
    \bottomrule[1.5pt]
    \end{tabular}%
    }
    \caption{\textbf{Effiectiveness as a universal Pixel-SSR paradigm.} All baselines are based on the unsupervised SSR, ``+ \textbf{Ours}" denotes training on our Pixel-SSR with 0.01\% label percentage, and Supervised denotes training on the image-level supervision. The best results of the three modes are \textbf{highlighted}. Our Pixel-SSR improves the results substantially than the unsupervised mode and achieves comparable results than the supervised image-level mode with significantly efficient label consumption. For clearer comparisons, the actual RMSE values are obtained by multiplying the values in the table by $10^{-2}$, for example, the RMSE of MST++ in CAVE is $ 4.20 \times 10^{-2} = 0.0420$.}
  \label{entire_paradigm}%
\end{table*}%

\begin{figure*}[h]
	\centering
	\includegraphics[width=0.95\textwidth]{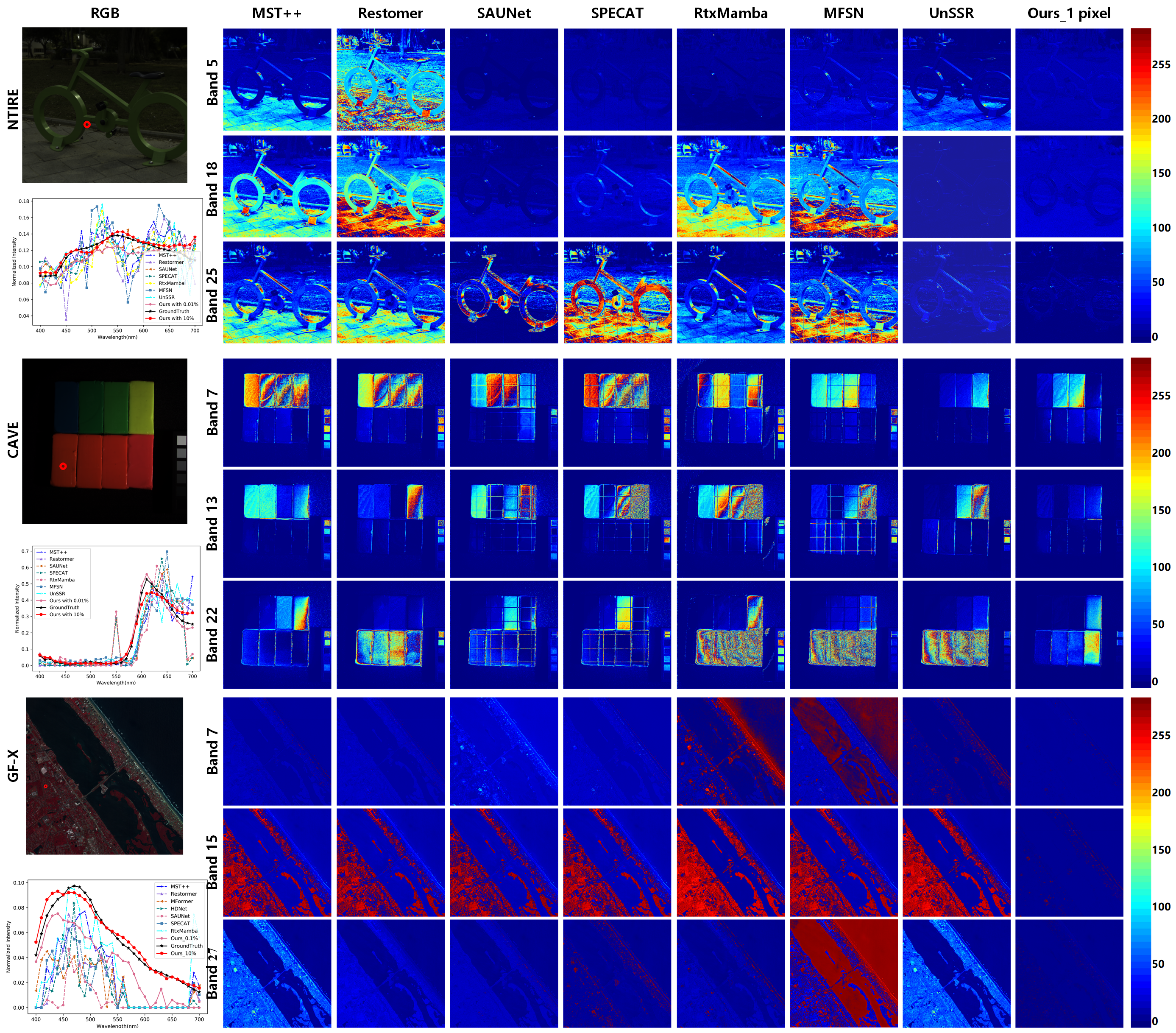}%
	\caption{ \textbf{Supplementary qualitative results of error maps.} Comparisons of the RMSE error maps in three spectral bands of the reconstructed HSI from Comparisons of the RMSE error maps of the reconstructed HSI from MST++/Restormer/SAUNet/SPECAT/RtxMamba/MFSN/UnSSR/Ours with 0.01\%. Besides, the spectral curves in a randomly selected pixel (marked as the red circle in RGB) are drawn in the bottom left. 
    Ours with 0.01\% label consumption consistently exhibits the darkest error maps and spectral curves most closely aligned with the groundtruth, signifying its superiority whether from a holistic or detailed aspect. 
    }
    \label{vis_sup_2020}
\end{figure*}

\begin{figure*}[h]
	\centering
	\includegraphics[width=0.95\textwidth]{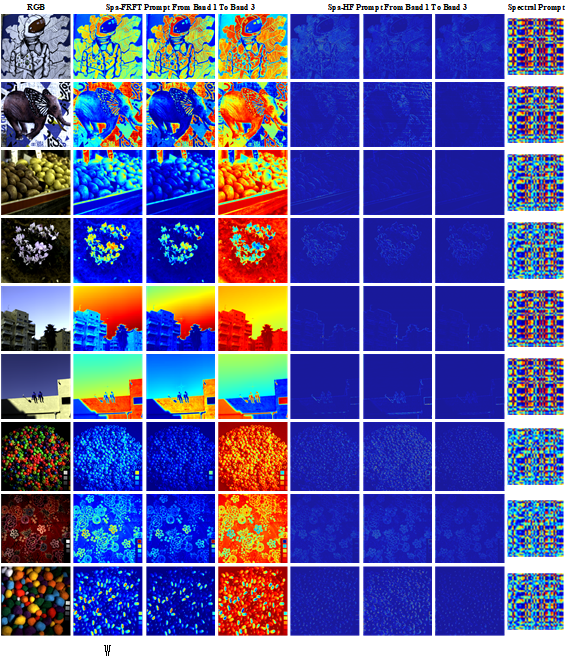}%
	\caption{ \textbf{Supplementary qualitative results of feature maps from DRPN.} In detail, Spa-FRFT Prompt $\mathcal{P}_{spa}$ reflects global spatial-wise content; Spa-HF Prompt $\mathcal{P}_{hf}$ delineates edge details; and Spectral Prompt $\mathcal{P}_{spe}$ denotes intrinsic spectral-wise dependency.
    }
    \label{vis_sup_drpn}
\end{figure*}

\section{More quantitative results on model stability and efficiency.}
\label{sup_com}
One of the challenging tasks for Pixel-SSR is how to implement the model in new scenes without point spectra.
We propose a Gamma-modeled strategy to synthesize point spectra during both training and inference, providing a fresh perspective to enhance the model's portability and robustness.
Here, we supplement from two aspects:
\begin{enumerate}
    \item{\textbf{Model Stability}: To rule out chance and objectively evaluate the stability of our method, we test each ratio 10 times and calculated the mean and variance of each evaluation criterion.}
	\item{\textbf{Model Efficiency}: To analyze the efficiency of Ours, the specific parameters, flops, and testing time consumption among Ours and other SOTA methods. }
\end{enumerate}

As calculated in Tables \ref{sup_junzhi_ntire}, \ref{sup_junzhi_cave} and \ref{sup_junzhi_gf}, the variances of all criteria in NTIRE, CAVE, GF-X are respectively below \textbf{$10^{-7}$}, \textbf{$10^{-10}$}, and \textbf{$10^{-8}$} orders of magnitude.
The results indicate that our method addresses the challenge of synthesizing point spectra for new scenes, effectively bridging this gap without inducing significant perturbations to the network's stability.

As listed in Table \ref{sup_params}, we further investigate the parameters, flops, and the testing time with a spatial size of $128 \times 128$ among all methods.
It can be indicated that Ours achieves the second-lowest Params and Flops, and an acceptable testing time due to the calculating consumption in the FRFT domain to depict abundant intrinsic characteristics.

\section{More quantitative results to verify the effectiveness of our Pixel-SSR.}
\label{paradigm}
To serve as a universal paradigm, we replaced the dotted box in Fig. \ref{framework} with other leading methods with the same loss constraints as our Pixel-SSR.
As shown in Table \ref{entire_paradigm}, all methods have been conducted in Our Pixel-SSR with 0.01\% label consumption to compare with those on unsupervised (without label) and supervised (with total entire images as the label) modes. 
All methods achieved significant improvements over the unsupervised mode and comparable results with the supervised mode, indicating the effectiveness of our Pixel-SSR as a universal paradigm.

\section{More ablation results.}
\label{sup_abla}

\subsection{Ablation on Gamma-modeled point spectra.}
As listed in Table \ref{sup_abla_gamma}, modeling point spectra using a Gaussian distribution resulted in different performance declines across all ratios. 
Thus, we deduce that gamma distribution is closer to the properties of the point spectra.

\subsection{Ablation results on NTIRE dataset.}
On the basis of which on CAVE in the manuscript, we perform the ablation study on NTIRE to investigate the components of our proposed DyPro-Mamba in Table \ref{sup_ala_ntire}.

\noindent{\textbf{Investigation of $\mathcal{P}_{spa}$ branch.}}
When we delete the spatial branch $\mathcal{P}_{spa}$, the model performance dropped 5.7851\%, 1.1044dB, 0.0012, and 8.0620\% on RMSE, PSNR, SSIM, and SAM. The results indicate that the intermediate space contains detailed non-stationary content, which depicts abundant global spatial-wise distributions that are vital to the reconstruction performance.

\noindent{\textbf{Investigation of $\mathcal{P}_{hf}$ branch.}}
When we remove the high-frequency representation $\mathcal{P}_{hf}$ branch, leaving $\mathcal{P}_{spa}$ and $\mathcal{P}_{spe}$. The lack of high-frequency components reduced the performance by 14.0496\%, 1.9084dB, 0.0062, and 19.945\% on RMSE, PSNR, SSIM, and SAM, showing the importance of edge details hidden in the high-frequency space.

\noindent{\textbf{Investigation of $\mathcal{P}_{spe}$ branch.}}
When we remove the spectral-wise dependancy $\mathcal{P}_{hf}$ branch, leaving $\mathcal{P}_{spa}$ and $\mathcal{P}_{hf}$. 
The performance decreased by 4.9587\%, 0.6179dB, and 9.2911\% on RMSE, PSNR, and SAM without the guidance of 
pre-extracted spectral guidance, indicating that the extracted spectral-wise dependency  is vital for the Point-SSR task.

\noindent{\textbf{Investigation of DRPN.}}
When deleting the whole prompting learning module DRPN, RMSE, PSNR, SSIM, and SAM respectively dropped by 15.7025\%, 2.0222dB, 0.0053, and 16.7691\%, which demonstrates that our DRPN improves the reconstruction performance via effectively mining available information from RGB and constrained point HSI.

\noindent{\textbf{Investigation of PromptSSM.}}
When removing PromptSSM, the accuracy decreased by 17.3554\%, 1.9899dB, 0.0099, and 22.2048\% on RMSE, PSNR, SSIM, and SAM, indicating the necessity of the guidance of spectral-wise dependency.

\section{More qualitative details.}
\label{sup_vis}
\noindent{\textbf{Error Maps.}} To rigorously evaluate the reconstruction performance of SOTA methods, we supplement performed RMSE heat maps across three random bands of all datasets, as shown in Fig. \ref{vis_sup_2020}. 
Additionally, spectral curves were plotted in a random pixel, highlighted as a red circle in the RGB image, with corresponding spectral profiles displayed for comparison. 
In the RMSE maps, darker colors indicate higher reconstruction accuracy. Our approach consistently exhibits the darkest error maps and spectral curves most closely aligned with the groundtruth, signifying its superiority whether from a holistic or detailed aspect. 

\noindent{\textbf{Feature Maps.}} As one of our core ideas, DRPN dynamically delineates complementary prompts: Spa-FRFT Prompt, Spa-HF Prompt, and Spectral Prompt, which direct the spatio-spectral reconstruction with global spatial distributions, edge details, and spectral dependency.
To further analyze our proposed DRPN, we demonstrate more feature maps of DRPN in Fig. \ref{vis_sup_drpn}:
\begin{itemize}
	\item $\mathcal{P}_{spa}$ depicts global spatial distributions, demonstrating 
     the capability to capture non-stationary content through Fractional Fourier Transform (FRFT) in a hybrid-sweep way.
    \item $\mathcal{P}_{hf}$ excavate the complementary spatial texture and edge clarity, indicating the ability to extract high-frequency details.
    \item $\mathcal{P}_{spe}$ delineates spectral correlations through self-attention, showing the importance of integrating outputs from $\mathcal{P}_{spa}$ and $\mathcal{P}_{hf}$ into the Gamma-modeled point spectra for intrinsic dependency.
\end{itemize}

\end{document}